\documentclass[conference]{IEEEtran}
\usepackage{times}

\usepackage[numbers]{natbib}
\usepackage{multicol}
\usepackage{graphicx}
\usepackage{amsmath}
\usepackage[bookmarks=true]{hyperref}

\begin{document}

\title{Fast Anticipatory Motion Planning for Close-Proximity Human-Robot Interaction}


\author{Sam Scheele\authorrefmark{1}, Pierce Howell\authorrefmark{1}, Harish Ravichandar\authorrefmark{1}
\\
\authorblockA{\authorrefmark{1} Structured Techniques for Algorithmic Robotics Lab, Georgia Institute of Technology, Atlanta GA 30318
\\ \texttt{\{scheele,pierce.howell ,harish.ravichandar\}}@gatech.edu}}



%

\maketitle

\begin{abstract}
Effective close-proximity human-robot interaction (CP-HRI) requires robots to be able to both efficiently perform tasks as well as adapt to human behavior and preferences. However, this ability is mediated by many, sometimes competing, aspects of interaction. We propose a real-time motion-planning framework for robotic manipulators that can simultaneously optimize a set of both task- and human-centric cost functions. To this end, we formulate a Nonlinear Model-Predictive Control (NMPC) problem with kino-dynamic constraints and efficiently solve it by leveraging recent advances in nonlinear trajectory optimization. We employ stochastic predictions of the human partner's trajectories in order to adapt the robot's nominal behavior in anticipation of its human partner. Our framework explicitly models and allows balancing of different task- and human-centric cost functions. While previous approaches to trajectory optimization for CP-HRI take anywhere from several seconds to a full minute to compute a trajectory, our approach is capable of computing one in 318 ms on average, enabling real-time implementation. 
We illustrate the effectiveness of our framework by simultaneously optimizing for separation distance, end-effector visibility, legibility, smoothness, and deviation from nominal behavior. We also demonstrate that our approach performs comparably to prior work in terms of the chosen cost functions, while significantly improving computational efficiency.
\end{abstract}

\IEEEpeerreviewmaketitle

\section{Introduction}
\subsection{Adaptive Close-Proximity HRC}
Adaptive close-proximity human-robot collaboration provides both psychological advantages to humans (in the form of increased comfort during interation) and efficiency advantages in task completion. Though the efficiency gains due to adaptive motion planning likely vary significantly by task, \citet{just1} found that adaptive, anticipatory motion planning improved task completion speed, reduced robot idle time, and increased separation distance relative to non-human-aware approaches on an assembly task.

A great deal of work in anticipatory motion planning focuses on non-collision, such as 
\citet{funnels} and \citet{MPCC}. 
\citet{legibility} notably introduced the concept of legibility, and found in \cite{just2} that legible motion leads to greater team fluency and closer collaboration, but their method for generating legible motion does not model humans as participatory actors, but rather as passive observers. 

\subsection{Nonlinear Model-Predictive Control}
If the factors associated with natural collaboration can be modeled as objective functions, a natural model for human-anticipatory motion planning is Model-Predictive Control (MPC, or NMPC for a nonlinear system model), in which we solve an optimization problem to find optimal control inputs based on predictions of system state for some time horizon. With NMPC, we can provide an arbitrary cost function and obtain a locally optimal solution.

Use of MPC for HRC is not new - \citet{faroni2019mpc} proposed an MPC-based approach that modified plans from a high-level planner to adaptively slow trajectories that get too close to humans and potentially adjust them to satisfy secondary objectives. \citet{yang2022fluidgrasp} used the GPU-accelerated stochastic MPC solver proposed by \citet{bhardwaj2022storm} to generate fluid motion for human-to-robot handovers. However, we are not aware of any other approaches that use MPC to realize anticipatory and reactive motion planning while optimizing the path itself for natural collaboration with humans.

\begin{figure}
    \centering
    \includegraphics[scale=0.4]{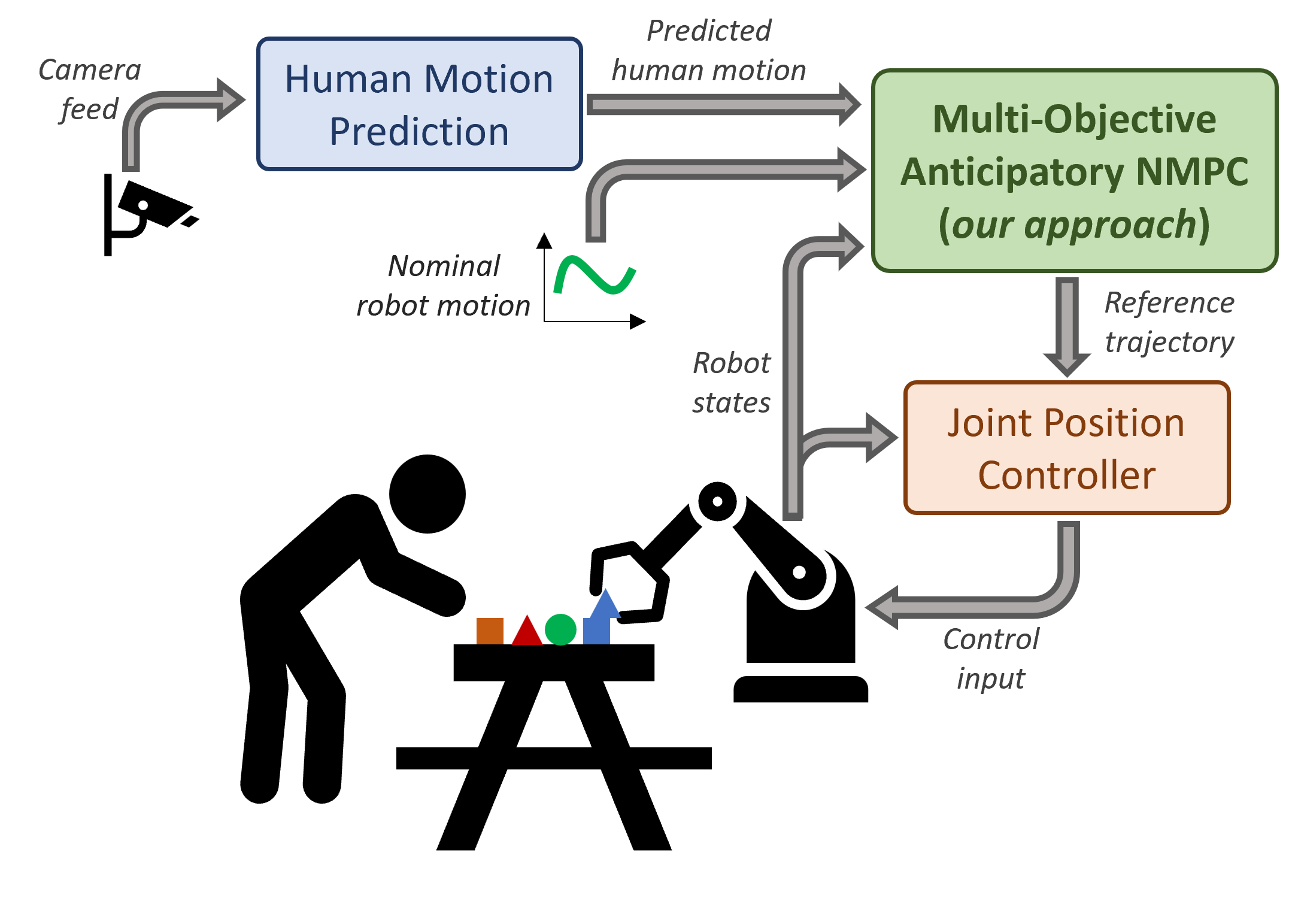}
    \caption{High-level flowchart of our method}
    \label{fig:my_label}
\end{figure}

\subsection{Our Approach}
In this work, we formulate human-anticipatory trajectory optimization as a NMPC problem and solve that problem using ALTRO \cite{altro}, a solver designed for nonlinear trajectory optimization. ALTRO uses a variant of the Iterated Linear-Quadratic Regulator (iLQR) algorithm to warm-start its solutions and achieve fast solution times, and it terminates when the performance ratio falls below a threshold. We use a sliding-window approach and implement optimizations such that the problem can be solved in real-time, enabling the system to be both anticipatory and reactive.

Relative to our previous work in multi-objective trajectory optimization, CoMOTO \cite{comoto_v1}, our approach offers the advantage of being more compatible with modern pose estimation: the proof-of-concept provided in CoMOTO relied on a single prediction of a full human trajectory, and took tens of seconds to optimize a solution. In contrast, our work relies only on predictions for short time horizons, which is consistent with human pose estimation works like \cite{PGBIG} and \cite{GSPS}, neither of which predict more than two seconds ahead. Further, our method is fast enough that it can run in real time, meaning that it can react to unexpected human motion.

\section{Method}

We apply NMPC to a manipulator equipped with joint-space position control, where the predictions for the MPC are generated by the human pose prediction network from \citet{GMMpred}. The problem is formulated as:

\begin{align}
\min_u \quad &C(X_{R}, X_{H}) \quad s.t.\\
X_R^{(0)} &= start \\
X_R^{(t+1)} &= X_R^{(t)} + u^{(t)} \Delta t\\
X_R^{(0)} &= goal \\
B_L \leq &u^{(t)} \leq B_U
\end{align}

Where $X_R$ and $X_H$ are respectively the robot (joint-space) and human (cartesian-space) trajectories, $C$ is a cost function to be presented below, $\Delta t$ is a predefined timestep (in our case, 0.2 sec), and $B_L$ and $B_U$ are limits on the joint velocity control vector $u$. This approach assumes that human movement is not reactive to robot movement. This assumption is likely false, but it may be approximately true over short timespans (violation of this assumption in the long-term might reasonably be handled by the reactive nature of our approach, albeit with some consequences for optimality). Note that while the MPC model assumes the robot is equipped with a joint-space velocity controller, in reality we use a joint-space position controller. The use of velocity control in the MPC model allows us access to a greater range of cost functions without sacrificing model accuracy.

In our prior work, we formulated an overall cost function as a weighted sum of five objectives. In this work, we reformulate several of these objectives to allow each timestep in the trajectory to be optimized independently of the others.

\subsection{Distance Cost}
The distance cost is the sum of pairwise distances between all human and robot joints, scaled by the human prediction covariance matrices:

$$ C_{dist}(x, u, t) = \sum_h \sum_r \dfrac{1}{d_{h,r}(t)^T (\Sigma_h(t))^{-1} d_{h,r}(t)} $$

Where $d_{h,r}(t)$ is the difference between the Cartesian positions of human joint $h$ and robot joint $r$ at time $t$ and $\Sigma_h(t)$ is the prediction covariance matrix of human joint $h$ at time $t$
\subsection{Visibility Cost}
The visibility cost remains the same as in CoMOTO: 

$$ C_{vis}(x, u, t) = \dfrac{\angle(O, \mu_{head}(t), p_{eef}(t))}{\sigma_{head}(t)} $$

Where $\angle(O, \mu_{head}(t), p_{eef}(t))$ is the angle between the object the person is assumed to be looking at, the estimated position of their head, and the robot's end effector, and $\sigma_{head}(t)$ is the standard deviation in the estimation of the human head position.

\subsection{Legibility Cost}
Legibility measures the ability of a human observer to predict the actual goal of an agent $G_R$ from a set of candidate goals $G$ (for example, an arm might reach for one of a few objects). Legibility can be naturally expressed in terms of probabilities and is defined by Dragan et. al. as

$$\dfrac{\int P(G_R | \xi_{S \rightarrow \xi(t)}) f(t) dt}{\int f(t) dt} $$

Where the probability term is given as:

$$ P(G_R | \xi_{S \rightarrow \xi(t)}) = \dfrac{1}{Z} \dfrac{\exp(-\mathcal{D}[\xi_{S \rightarrow Q}] - V_{G_R}(Q))}{\exp(-V_{G_R}(S))} $$

Here, $\xi$ is the , $S$ is the trajectory's start point ($\xi_0 = S$), $V_{G_R}(Q)$ is a function that gives the minimum value of $C$ for any trajectory from $Q$ to $G_R$, and .

Where:

\begin{itemize}
    \item $\xi$ is the trajectory of a robot (in our case, the cartesian-space trajectory of the end effector)
    \item $f(t)$ is a weighting function that allows different portions of the trajectory to receive different weight
	\item $Z$ is a normalizer ($Z = \sum_G  \dfrac{\exp(-\mathcal{D}[\xi_{S \rightarrow Q}] - V_{G}(Q))}{\exp(-V_{G}(S))}$)
    \item $S$ and $Q$ are the starting and current positions, respectively ($Q=\xi(t)$)
    \item $G_R$ is the robot's end-effector goal
    \item $\mathcal{D}$ is a cost functional, defined for our code as the square of the path length
    \item $V_A(X)$ is defined as the cost of the lowest-cost path from $X$ to $A$ ($\texttt{min}_\xi \mathcal{D}[\xi] | \xi(0) = X, \xi(end) = A$). 
\end{itemize}
  
The formulas above are implemented in CoMOTO using a summation to approximate the integral and piecewise distances between timesteps to approximate path lengths for $\mathcal{D}$. CoMOTO expressed the legibility cost at a timestep $t$ as:

$$ \textsc{LegibilityCost}[\xi, t] = 1 - P(G_R | \xi_{S \rightarrow \xi(t)})  $$

The CoMOTO formulation for $ P(G_R | \xi_{S \rightarrow \xi(t)})$ couples the trajectory waypoints, making optimization more difficult. To enable optimization of each waypoint independently, we decouple the waypoints by modifying probability function:

$$ P(G_R | \xi_{S \rightarrow Q}) = \dfrac{1}{Z} \dfrac{\exp(-\hat{\mathcal{D}}(S, Q, t) -||G_R - Q||^2)}{\exp(-||G_R - S||^2)} $$

The first change to the function has been to define $\mathcal{D}(\xi) = ||\xi||^2$, then substitute $V_{G_R}$ for its value under our definition of $\mathcal{D}$, $V_{G_R}(Q) = ||G_R - Q||^2$. Second, we define $\hat{\mathcal{D}}(S, Q, t)$, an approximation of $\mathcal{D}(\xi_{S \rightarrow Q})$ parameterized only on $S$, the starting point, $Q$, the end of the trajectory segment, and $t$, the time (or, equivalently, knotpoint index) at which the trajectory segment ends ($\xi_t = Q$). The motivation for this will be clear momentarily.

Implementing the normalization constant $Z$ similarly to $P(G_R | \xi_{S \rightarrow Q})$:

$$ Z = \sum_G  \dfrac{\exp(-\hat{\mathcal{D}}(S, Q, t) - ||G - Q||^2)}{\exp(-||S - G||^2)} $$

We observe that the $\hat{\mathcal{D}}(S, Q, t)$ term is not parameterized on $G$ and can therefore be factored out of the sum, so the $\exp(-\hat{\mathcal{D}}(S, Q, t))$ term will cancel with the one in the definition of $P(G_R | \xi_{S \rightarrow \xi(t)})$. Therefore, the legibility cost can also be written as:

$$ C_{leg}(x, u, t) = 1 - \dfrac{1}{Z} \exp(||G_R - S||^2 - ||G_R - x||^2)  $$

\subsection{Nominal Cost}
The nominal cost also remains the same as the CoMOTO cost; it is defined at each waypoint as the cartesian distance between the nominal and actual end-effector trajectories:

$$ C_{nom}(x, u, t) = ||p_{eef}(t) - p_{eef}^*(t)|| $$

\subsection{Smoothness Cost}
The smoothness cost is defined as the squared magnitude of joint-space velocity at each timestep:

$$ C_{smooth}(x, u, t) = ||u||^2 $$

\subsection{Goal Pose Cost}
The goal pose cost is designed to incentivize the MPC solutions to approach the goal (which it does in conjunction with the nominal cost). It is defined as the distance between the end effector's cartsian position at the goal pose at a given waypoint, plus an orientation error term:

$$ C_{goal}(x, u, t) = ||p_{goal} - p_{eef}|| + 1 - \langle q_{goal}, q_{eef} \rangle^2$$

Where $q_{eef}$ and $q_{goal}$ are unit quaternions giving the orientation of the end effector at the current time and goal position, respectively.

\subsection{Combined Cost and Problem Setup}
The costs presented above are combined in a weighted sum to yield the optimization objective $C$ from equation 1, where the weights were tuned by hand to produce subjectively natural trajectories. Nonlinear solvers often require an initial solution in the feasible region of the problem as a starting point; our work uses a joint-space linear trajectory towards the (joint-space) goal as the initial solution.

\section{Experiments and Metrics}
To validate our new approach against CoMOTO, we applied each approach to a set of predicted human trajectories made by the GMM from \citet{GMMpred} and associated covariances in a reaching task, where the robot had to reach across a human to reach a desired pose. Each trajectory lasted 5 seconds, and the NMPC was configured with a 1.25 second time horizon and replanned every 0.5 seconds. Both our method and CoMOTO were configured with a timestep of 0.25 seconds. Since we expected the approximations and moving horizon in our approach to result in slightly suboptimal trajectories, we hypothesized that our approach would perform slightly worse than CoMOTO, but run significantly faster.

Our new approach is written in Julia, which is a just-in-time compiled language. Therefore, to avoid capturing the time to JIT compile our approach in our metrics, we solved the optimization problem once and discarded the results before solving it again to measure the wall clock solution time. To ensure that this didn't provide our approach with any advantages in terms of cache availability, we also ran CoMOTO once before testing (CoMOTO is written in Python and C++ and so doesn't have a JIT compiler).  We evaluate on the following metrics:

\textbf{Separation distance (Dst.)}: the percentage of the trajectory where the distance between the human and any part of the robot exceeds 20cm.

\textbf{End effector visibility (Vis.)}: the percentage of the trajectory where the robot's end effector is within the human's field of view, assuming they're looking at the object they're reaching for.

\textbf{Legbility (Leg.)}: average $P(G_R | \xi_{S \rightarrow Q})$, as given by \citet{legibility}.

\textbf{Deviation from nominal trajectory (Nom.)}: the sum of square distances between corresponding timesteps of the nominal and computed end effector trajectories.

\textbf{Latency (Lat.)}: the average time required to obtain a robot trajectory once a human trajectory has been predicted

The first four of these metrics are the same as those chosen for the CoMOTO paper - scoring higher is better on all metrics except the last two (latency and deviation from nominal). Higher deviation from the nominal trajectory isn't universally better or worse than lower deviation, but, all else being equal, we take lower deviation to be better.

\section{Results}

\begin{table}[]
    \centering
    \begin{tabular}{c|c|c}
        & Ours & CoMOTO \\
        \hline
         Dst. & \textbf{0.83} (0.17) & 0.52 (0.47) \\
         Vis. & 0.53 (0.01) & \textbf{0.74} (0.03)\\
         Leg. & 0.26 (0.01) & \textbf{0.60} (0.01)\\
         Nom. & \textbf{5.80} (0.67) & 9.79 (1.64)\\
         Lat. (sec) & \textbf{0.318} (0.027) & 125.73 (27.22)
    \end{tabular}
    \caption{Performance of our method vs CoMOTO. Best performances are bolded, and standard deviations are parenthesized.}
    \label{tab:perf_stats}
\end{table}

\begin{figure}
    \centering
    \includegraphics[scale=0.2]{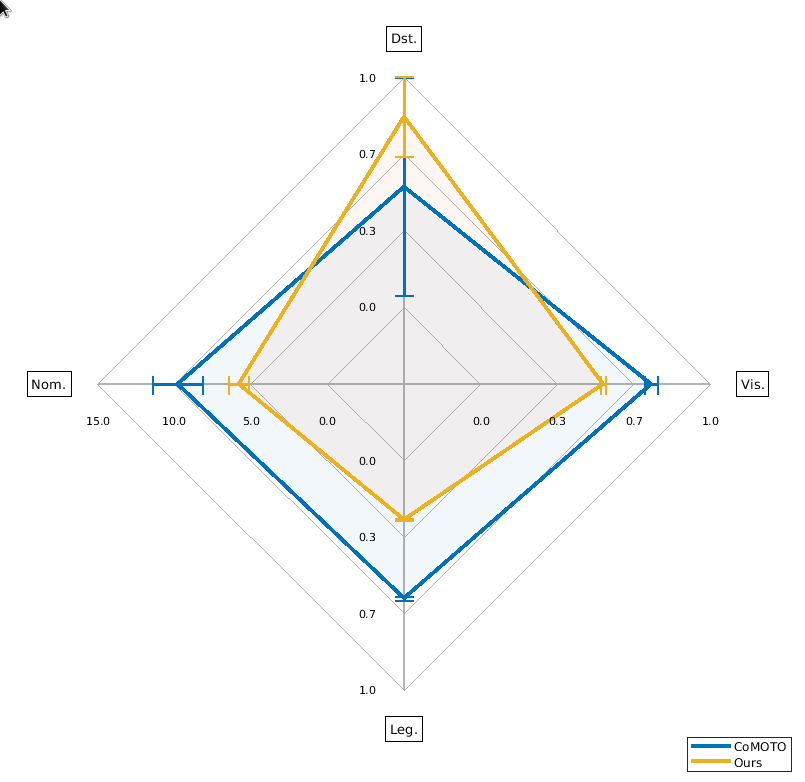}
    \caption{Performance of our method vs CoMOTO. Higher metrics are better, except on nominal (Nom.)}
    \label{fig:spider-plot}
\end{figure}

Our results are summarized in table \ref{tab:perf_stats} and figure \ref{fig:spider-plot}. As expected, CoMOTO performs better than our approach on two out of three metrics. However, our approach not only performs better on the distance metric (which is most important for human safety), but it is more than 400 times faster than CoMOTO. Further, our approach has lower standard deviations than CoMOTO, meaning that it is more reliable. This is an unexpected result, as our approach does not have access to the entire trajectory prediction ahead of time, and is therefore capable of making planning errors that CoMOTO would avoid. 

Unexpectedly, our method outperforms CoMOTO on the distance metric while underperforming it on legibility and visibility. We believe that some of this variation is due to differences in the cost functions used by our method. Visually, our method produces trajectories similar to CoMOTO's, but with less deviation from nominal behavior.

\section{Conclusion} 
In this paper, we present an MPC-based approach to human-anticipatory trajectory optimization. We formulate several cost functions which may be relevant to natural HRC and use our approach to optimize for a weighted combination of these costs, balancing them against each other. While prior work required tens of seconds to solve this problem, our approach solves it in about 300ms, which approaches real-time speeds and enables far broader application than offline approaches.

In the future, we believe that exploring how humans move in reaction to the movements of their robot partners may be a useful area of study. Additionally, new solvers or formulation approaches may further reduce the solution latency, improve quality, or provide stronger theoretical safety guarantees.

\bibliographystyle{plainnat}
\bibliography{references}

\end{document}